\documentclass[10pt,twocolumn,letterpaper]{article}

\usepackage{cvpr}              % To produce the CAMERA-READY version
\usepackage{multirow}
\usepackage{color}
\usepackage{stfloats}
\definecolor{cvprblue}{rgb}{0.21,0.49,0.74}
\usepackage[pagebackref,breaklinks,colorlinks,allcolors=cvprblue]
{hyperref}

\def\paperID{11027} % *** Enter the Paper ID here
\def\confName{CVPR}
\def\confYear{2025}

\title{Learning Bijective Surface Parameterization for Inferring Signed Distance Functions from Sparse Point Clouds with Grid Deformation}
\author{%
{Takeshi Noda}$^1$\thanks{Equal contribution. $^\dagger$The corresponding author is Yu-Shen Liu. This work was partially supported by Deep Earth Probe and Mineral Resources Exploration ­—  National Science and Technology Major Project (2024ZD1003405), and the National Natural Science Foundation of China (62272263).} \ \quad {Chao Chen}$^1$$^*$ \ \quad {Junsheng Zhou}$^1$ \quad {Weiqi Zhang}$^1$ \ \quad \ \\ 
\quad {Yu-Shen Liu}$^1$$^\dagger$ \ \quad {Zhizhong Han}$^2$\     \\
  $^1$School of Software, Tsinghua University, Beijing, China \\
  $^2$Department of Computer Science, Wayne State University, Detroit, USA \\
  \tt\small{\{yeth24,zhou-js24,zwq23\}@mails.tsinghua.edu.cn}\quad \\
  \tt\small{chenchao19@tsinghua.org.cn}\quad
  \tt\small{liuyushen@tsinghua.edu.cn} \quad 
  \tt\small{h312h@wayne.edu}
  }
\begin{document}
\maketitle

\begin{abstract}
Inferring signed distance functions (SDFs) from sparse point clouds remains a challenge in surface reconstruction. The key lies in the lack of detailed geometric information in sparse point clouds, which is essential for learning a continuous field. To resolve this issue, we present a novel approach that learns a dynamic deformation network to predict SDFs in an end-to-end manner. To parameterize a continuous surface from sparse points, we propose a bijective surface parameterization (BSP) that learns the global shape from local patches. Specifically, we construct a bijective mapping for sparse points from the parametric domain to 3D local patches, integrating patches into the global surface. Meanwhile, we introduce grid deformation optimization (GDO) into the surface approximation to optimize the deformation of grid points and further refine the parametric surfaces. Experimental results on synthetic and real scanned datasets demonstrate that our method significantly outperforms the current state-of-the-art methods. Project page: \url{https://takeshie.github.io/Bijective-SDF}
%, demonstrating its great potential for sparse point cloud reconstruction tasks.
\end{abstract}    
\vspace{-0.2cm}
\section{Introduction}
\label{sec:intro}

Surface reconstruction from 3D point clouds is an important task in 3D computer vision. Continuous surfaces are widely used in downstream applications, such as autonomous driving, VR, and robotics. With the development of deep learning \cite{atzmon2020sald,zhou20223d,genova2019learning,DBLP:conf/icml/GroppYHAL20,tang2021sa,zhou2023uni3d,Zhou2023VP2P,li2023neaf}, significant breakthroughs have been made in learning signed distance functions (SDFs) to represent continuous surfaces \cite{jiang2020local,duan2020curriculum}. The SDFs learned from dense point clouds are continuous and complete, which allow us to obtain robust isosurfaces of discrete scalar fields using the Marching-cubes algorithm \cite{lorensen1987marching}. However, when confronted with sparse point clouds, current approaches fail to accurately predict a signed distance field around the surface, impacting their performance on the real-world scenario where only sparse point clouds are available.

Previous works \cite{erler2020points2surf,huang2022neural,mi2020ssrnet,VisCovolume} which infer the SDFs from raw point clouds typically
require ground truth signed distances or dense point clouds as supervision. With sparse point clouds, current state-of-the-art methods \cite{On-SurfacePriors} obtain shape priors from large scale supervision to handle the sparsity. Although prior-based methods can leverage data-driven information to infer SDFs with simple topological structures, they still struggle to  deal with real-world diverse sparse inputs containing complex geometries. Some methods learn self-supervised priors from sparse point clouds to maintain the shape integrity \cite{chen2023unsupervised,ouasfi2024unsupervised}. However, these approaches still do not work well with sparse points, making it remain challenging to recover the complete geometries.

To address this issue, we propose Bijective Surface Parameterization (BSP) for learning a continuous global surface. Unlike previous approaches, we construct a continuous bijective mapping between the canonical spherical parametric domain and the 3D space. For each 3D point, we transform it into a code on the sphere manifold in the parameter space, and then regard the code as a center to densify the patch it locates by sampling more codes on the sphere. With the learned BSP, we transform each densified patch in the parameter space back into 3D space, leading to a 3D shape with denser points. 
Compared to methods which directly upsample a global shape, we train local patches with a shared network to recover more details on patches.
 
Based on the densified points, we propose a Grid Deformation Optimization (GDO) strategy to estimate the SDFs. Our key insight is to utilize deformable tetrahedral grids to generate watertight shapes under the supervision of densified points. To this end, we sample uniformly distributed vertices from the generated shape to 
match the densified points. It allows us to progressively learn SDFs from coarse to fine.
Extensive experiments on widely-used benchmark datasets demonstrate that our method significantly outperforms the current state-of-the-art methods. Our contributions are summarized as follows:

\begin{itemize}
\item We propose a novel framework that learns the neural deformation network to infer signed distance fields from sparse points without additional surface priors.
\item We demonstrate that learning bijective surface parameterization can parameterize the surface represented by sparse points, which introduces a novel way of sampling dense patches in the parameter space.
\item We achieve state-of-the-art results in surface reconstruction on synthetic data and real scenes in widely used benchmarks, demonstrating the great potential in sparse reconstruction tasks.
\end{itemize}

\section{Related Work}
\label{sec:relatedwork}
Surface reconstruction from 3D point clouds has made significant progress over the years \cite{tang2021sa,chabra2020deep,liu2021deep,lombardi2020scalable,martel2021acorn,zhou2024fast,zhou2024cap,Zhou2022CAP-UDF}. Earlier optimization based methods infer continuous surfaces from point clouds. With the development of datasets \cite{shapeneturl,geiger2012we}, deep learning methods achieved promising results. In particular, learning the neural implicit field (NIF) has been widely applied in various reconstruction tasks, including multi-view reconstruction \cite{mildenhall2020nerf,johari2022geonerf,zhang2023fast,zhang2024learning,zhang2024gspull,zhang2025nerfprior,zhang2025monoinstance,li2025gaussianudf,AniSDF,han2024binocular,zhou2024diffgs,huang2023neusurf}, point cloud reconstruction \cite{guerrero2018pcpnet,PredictiveContextPriors,yifan2021iso,wen20223d,li2024LDI}, and occupancy estimation \cite{mescheder2019occupancy,peng2020convolutional,mi2020ssrnet,zhou2024deep}. In the following section, we focus on implicit representation learning methods based on sparse point clouds.\\
\textbf{Neural Implicit Surface Reconstruction.} In recent years, a lot of advances have been made in 3D surface reconstruction tasks with NIF methods. NIF represents shapes in implicit functions using occupancy  \cite{tang2021sa,peng2020convolutional,chen2019learning} or signed distance functions  \cite{ma2023towards,ma2023learning,noda2024multipull,zhou2023levelset,udiff} and reconstructs surfaces with the Marching-cubes algorithm. Previous studies employ the global optimization based strategy  \cite{ma2021neural,sitzmann2020implicit}, embedding objects as latent codes to predict the NIF. Furthermore, to reconstruct finer local details, some methods use different training strategies to capture the local priors  \cite{tretschk2020patchnets,chibane2020implicit}. In addition, some recent methods introduce new perspectives for learning NIF through differentiable Poisson solvers  \cite{peng2021shape}, iso-points  \cite{yifan2021iso}, and grid interpolation \cite{chen2023gridpull}. However, these methods rely on dense point cloud inputs or signed distances and normals, which limits their ability to accurately predict NIF from sparse point clouds.\\
\textbf{Learning Self-Priors from Sparse Points Clouds.}
Learning NIF from sparse point clouds without ground truth as supervision is a more intricate task. Onsurf  \cite{On-SurfacePriors} manages to understand sparse points by using pretrained priors. However, it is limited by the weak generalization ability for diverse inputs. Some studies focus on learning NIF from sparse point clouds. Ndrop \cite{boulch2021needrop} introduces a statistical strategy to learn the decision function for implicit occupancy fields with sample points. However, this method struggles to constrain all sample points accurately. Building on this, SparseOcc \cite{ouasfi2024unsupervised} utilizes a classifier to simplify the process of learning occupancy functions, which significantly improves the efficiency to learn occupancy field from sparse inputs. Despite these advancements, these methods only depend on sparse inputs as supervision, the learned decision functions tend to produce coarse approximations and fail to handle extremely sparse or complex inputs. Meanwhile, inferring smooth surfaces from occupancy fields is still challenging. VIPSS \cite{huang2019variational} and SparseOcc attempt to address this issue through Hermite interpolation and entropy-based regularization. However, these approaches are sensitive to hyper-parameters and lack general applicability.\\
\textbf{Learning Parametric Surfaces from Sparse Point Clouds.}
Previous studies \cite{vakalopoulou2018atlasnet,yu2022part} proved that learning surface parameterization can map across dimensions and naturally infer the geometry of shapes. Although TPS \cite{chen2023unsupervised} explores its application to sparse point clouds, it is limited by sparse inputs and only learn parametric surface in global manner. TPS++ \cite{chen2024neuraltps} additionally introduces structure-aware distance constraints to enhance accuracy. However, it still struggles to learn the global geometry from local parametric surfaces. To address this issue, we propose bijective surface parameterization, which enables networks to learn multiple local parametric surfaces from the sparse point cloud and infer a finer global surface.\\
\textbf{Dynamic Deformation Network.} Neural deformation network  \cite{zou2024triplane,shen2023flexible,shen2021deep} dynamically learns 3D shapes with arbitrary topological structures using differentiable mesh vertex grids. GET3D  \cite{gao2022get3d} explicitly learns 3D models through a differentiable decoder to obtain detailed 3D models. By contrast, DynoSurf  \cite{yao2025dynosurf} learns keyframe point clouds as templates and uses neural networks to predict movement steps to obtain time-series 3D models with arbitrary topological structures. However, these methods rely on dense point clouds or pretrained embeddings. Here, we explore the feasibility of learning a deformation network for sparse tasks from a new perspective. During training, we learn parametric surfaces from sparse point clouds as supervision and learn implicit fields %through the deformation of tetrahedral vertices%
in a differential manner.
\begin{figure*}[!t]
  \centering
  \includegraphics[width=0.85\linewidth]{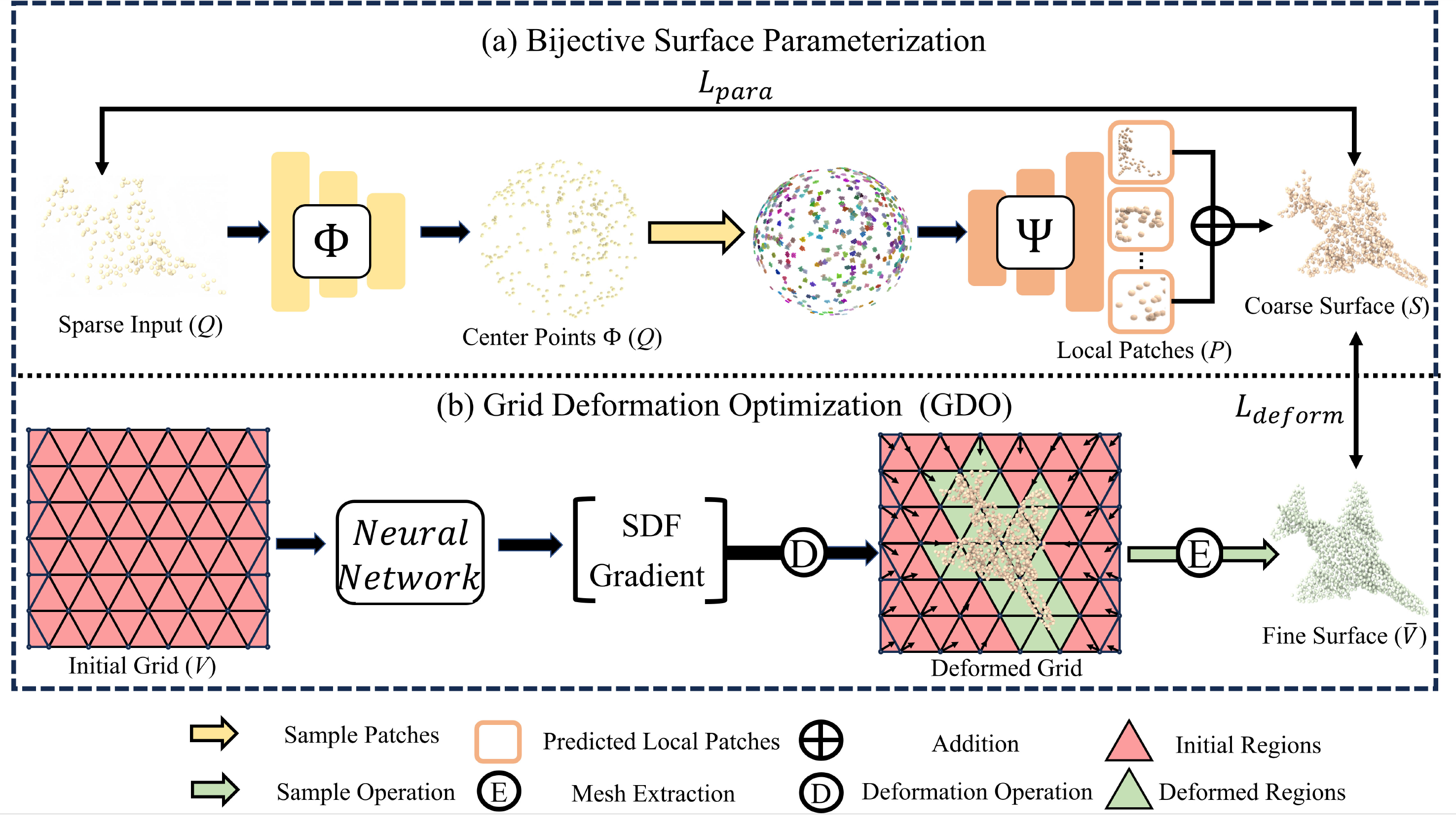}
  \vspace{-0.3cm}
  \caption{Overview of Our method. Given a sparse point cloud \( Q \), we first learn a mapping function \( \Phi \) to encode \( Q \) into a unit sphere parametric domain. We consider each point as center point and sample local patches on the parametric surface. Next, we learn the inverse mapping \( \Psi \) to predict the positions of these local patches in 3D space and integrate them to obtain \( S \). We leverage \( S \) as the supervision for the grid deformation network \( g \) and predict the signed distance field through the GDO optimization strategy. We further extract dense point cloud $\bar{V}$ from the implicit field and optimize the parameterized surface $S$.
}
  \label{fig:main_pic}
  \vspace*{-0.5cm}
\end{figure*}

\section{Method}
\label{sec:method}
\subsection{Overview}

Given a sparse point cloud $Q= \{ q_n \}_{n=1}^N$, we aim to learn a signed distance field that represents a continuous surface from $Q$. An overview of the proposed method is shown in Fig. \ref{fig:main_pic}. We present the bijective surface parameterization (BSP) in Fig. \ref{fig:main_pic} (a) to learn a continuous parametric surface representation. We first learn a canonical mapping  $\Phi$ to encode $Q$ into a unit sphere parametric domain $\mathcal{U}$, where we can sample local patches $P= \{ p_m \}_{m=1}^M$ around each point. Subsequently, we learn an inverse mapping $\Psi$ to decode $P$ back to 3D space and integrate patches into a global surface $S= \{ s_i \}_{i=1}^I$. With $S$ as supervision, we employ the grid deformation optimization (GDO) strategy to move the deformable grid points $ V= \{ v_j \}_{j=1}^J$ towards $S$ to infer the SDFs shown in Fig. \ref{fig:main_pic} (b). Our target is to minimize the differences between \( S \) and \( Q \). 
Moreover, we regulate the deformation of $V$ by constraining the distance between $V$ and $S$ to infer a continuous surface. To predict accurate SDFs, we also encourage $V$ to be on the zero level set of the field.
In this section, we will begin with introducing the bijective surface parameterisation (BSP). Subsequently, we will describe the grid deformation optimisation (GDO) strategy and loss functions in the following.
 
\subsection{Bijective Surface Parameterization}
Previous methods \cite{chen2023unsupervised} are constrained in representing continuous surfaces with multiple patches due to sparsity, which limits the completeness of the parametric surface. In contrast, we learn two mapping functions to achieve this: $\Phi$ maps each $q \in Q$ to the canonical parametric domain while $\Psi$ conducts the inverse mapping. We learn two mapping functions with an auto-encoder structure as follows.\\
\textbf{Canonical Mapping $\Phi$.} For each point $q_n \in Q$, we first encode the point-wise feature $\Phi(q_n)$ based on the PointTransformer \cite{vaswani2017attention} layer $\Phi$, which can be formulated as
\begin{equation}
\Phi(q_n) = \sum_{q_k=1}^{k} \rho \big(\gamma (\beta(q_n) - \eta(q_k) + \xi) \big) \odot (\alpha(q_k) + \xi),
\end{equation}
where $k$ is the set of k-nearest neighbors (KNN) of $q_n$, we set $k=8$ by default. $\{\alpha,\beta,\eta\}$ are linear layers, $\{\gamma$,$\delta\}$ are MLP layers, $\rho$ is the softmax function, position embeding $\xi=\delta(q_n-q_k)$, and $\odot$ is point-wise product operation.

\begin{figure}[!h]
  \centering
  \includegraphics[width=0.8\linewidth]{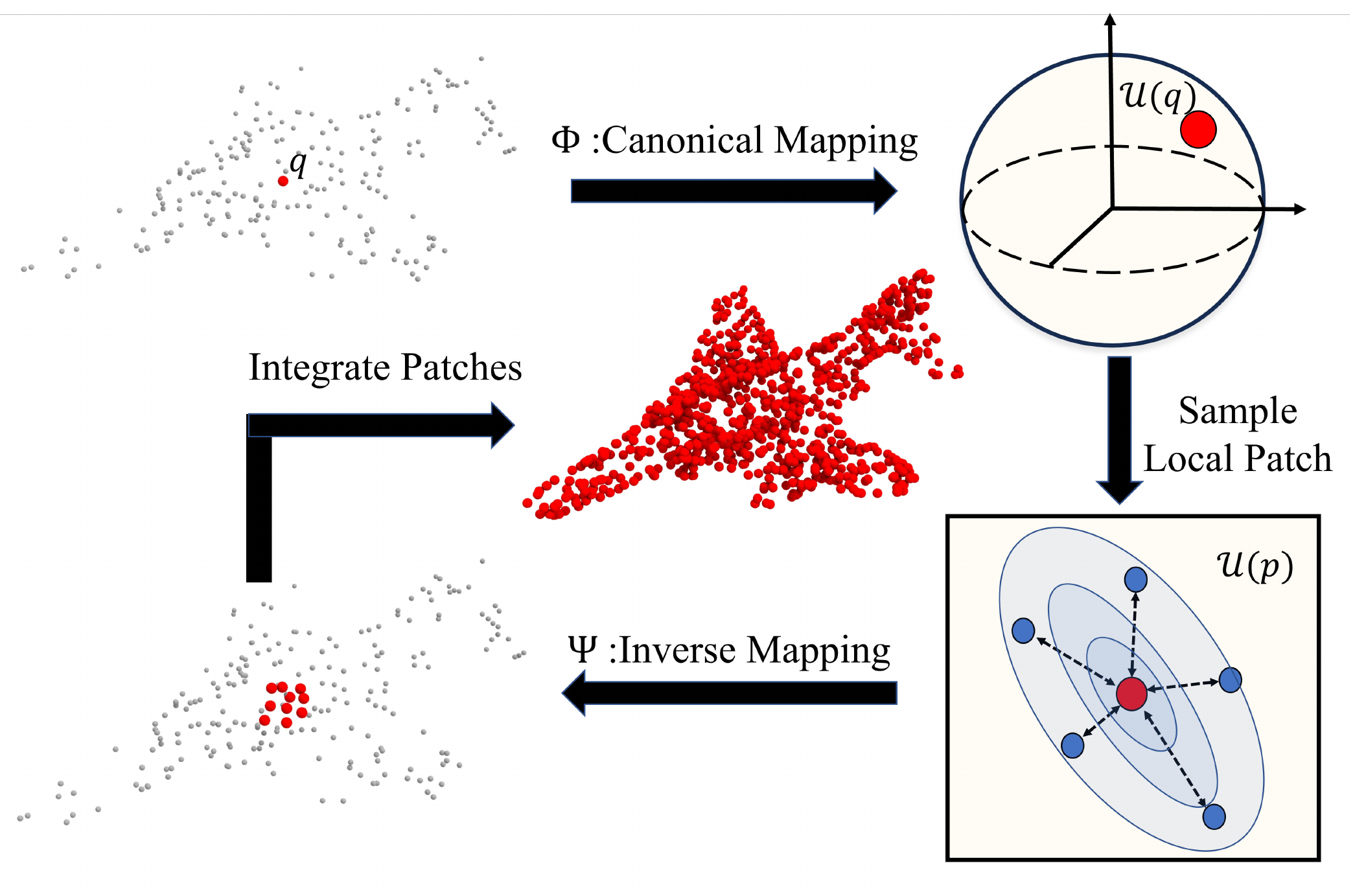}
  \caption{Illustration of BSP. The white points indicate the sparse input $Q$. For each point $q \in Q$, we learn mapping function \(\Phi\) to map $q$ to $\mathcal{U}(q)$, and sample a local patch $\mathcal{U}(p)$ on the parametric surface. Subsequently, we employ an inverse mapping \(\Psi\) to  assembles these patches into a global surface (red points).}
  \label{fig:bsp}
  \vspace*{-0.5cm}
\end{figure}

With the learned $\Phi$, we extract features into the coordinates which project them in the canonical unit sphere \(\mathcal{U}\), where \(\mathcal{U}(q_n) \in \mathcal{U}(Q)\). Each $\mathcal{U}(q_n)$ is a center point where KNN is utilized to sample a local patch $\mathcal{U}(p_m)$ around it. Specifically, we construct a uniform sphere coplanar with $\mathcal{U}(Q)$ to provide samples for $\mathcal{U}(q_n)$.\\
\textbf{Inverse Mapping $\Psi$.} Similarly, we efficiently estimate $\Psi \approx \Phi^{-1}$ with an neural network. We utilize the standard transformer decoder block as $\Psi$, which regards point-wise features with several linear layers $\varphi$ as the global condition, and local patch $\mathcal{U}(p_m)$ as queries to integrate a global shape $S$ in 3D space, which can be formulated as, 
\begin{equation}
\begin{aligned}
S = \sum_{m=1}^{M}\Psi(\varphi(\Phi(Q),\mathcal{U}(p_m))).
\end{aligned}
\end{equation}
We measure the distance between the parameterized surface $S$ and the sparse point cloud $Q$ using the Chamfer Distance (CD), denoted as $L_{para}$,
\begin{equation}
\begin{aligned}
L_{para} = \frac{1}{I} \sum_{s \in S} \min_{q \in Q} \|s - q\|^2 + \frac{1}{N} \sum_{q \in Q} \min_{s \in S} \|q - s\|^2.
\end{aligned}
\end{equation}
We visualize the BSP process in Fig. \ref{fig:bsp}. For \( q_n \in Q \), \(\Phi\) maps its position in \( \mathcal{U} \) and samples local patches $\mathcal{U}(p_m)$. With the inverse mapping \(\Psi\), we generate a local surface on \( S \). We integrate all local patches to obtain the global shape as the coarse surface $S$.
\subsection{Grid Deformation Optimization}
With the learned BSP, we parameterize the coarse surface $S$. Naive implementations rely on $S$ to infer SDFs and reconstruct surfaces, often producing holes due to the non-uniformity distribution. Unlike these methods, we design the grid deformation optimization strategy to learn continuous signed distance functions and further optimize parametric surfaces. Given tetrahedral grid points $V$, a straightforward strategy to update the deformed points $V'$ is to learn an offset $\varepsilon$ from neural network $g$, formulated as $V'=V + \varepsilon$.
However, directly learning offsets from $g$ fails to maintain consistency of deformation direction, resulting in difficulties in convergence. We maintain the consistency of the deformation by constraining on normals $n_V$ with gradients $\nabla g(V)$. During training, we predict the SDFs $g(V;\theta)$ and the gradient $\nabla g(V)$ to guide the deformation process of $V$. We consider $g(V)$  and $n_V$ to be the stride and direction, respectively. Therefore, the deformation process of $V$ can be described as 
\begin{equation}
\begin{aligned}
V \to V'=\|g(V;\theta)\cdot n_V-V\|_2,
\end{aligned}
\end{equation}
where $\theta$ is learnable parameter in deformation network $g$, $n_V=\ g(V;\theta)/\|\nabla g(V)\|_2$.
% \begin{equation}
% \begin{aligned}

% \end{aligned}
% \end{equation}

\begin{figure}[!h]
  \centering
  \includegraphics[width=0.9\linewidth]{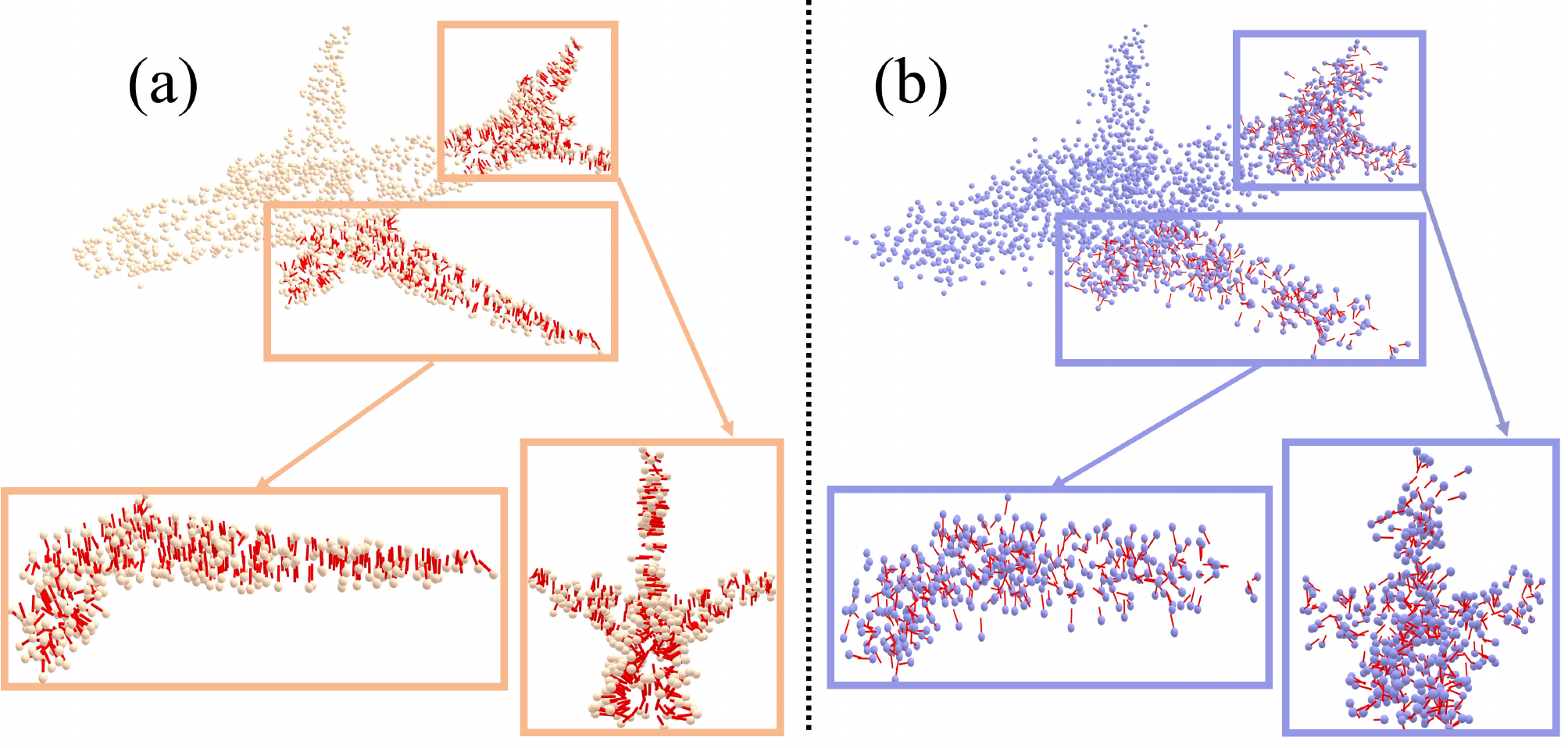}
  \caption{Visual comparison of GDO (a) and direct offset optimization (b), the red lines indicate the offset direction.}
  \label{fig:gdo}
  \vspace*{-0.3cm}
\end{figure}

We further compare the movement directions and optimization results of GDO in Fig. \ref{fig:gdo} (a) and the classical strategy \cite{shen2021deep} in Fig. \ref{fig:gdo} (b). The red lines indicate the next deformation direction of the grid points. Compared to direct offset prediction, GDO achieves more consistent deformation directions, resulting in a more accurate shape distribution.
Meanwhile, we extract the surface using Deep Marching Tetrahedra (DMT), the operation denoted as DMT$(\cdot)$.
The deformation grid points $\bar{V}$ on the surface can be expressed as $\bar{V}=$ DMT$(V')$, where $\bar{V}=\{\bar{v}_t\}_{t=1}^T$. We use the Chamfer Distance to regulate the deformation process of $\bar{V}$ and minimize the difference to $S$, denoted as $L_{deform}$, we have:
\begin{equation}
\begin{aligned}
L_{deform} = \frac{1}{T} \sum_{\bar{v} \in \bar{V}} \min_{s \in S} \|\bar{v} - s\|^2 + \frac{1}{I} \sum_{s \in S} \min_{\bar{v} \in \bar{V}} \|s - \bar{v}\|^2.
\end{aligned}
\end{equation}
To make the implicit field more accurate, we add the $L_{surf}$ term to encourage the network to learn zero level set from $g (V)$. Formulated as:
\begin{equation}
\begin{aligned}
L_{surf} = \mid g(V)\mid.
\end{aligned}
\end{equation}
Therefore, the total loss $L$ is given as:
\begin{equation}
\begin{aligned}
L =\lambda_{1}L_{para}+\lambda_{2}L_{surf}+L_{deform}, 
\end{aligned}
\end{equation}
where $\lambda_{1}$ and $\lambda_{2}$ are weight parameters, which we set to 10 and 0.01 by default.
\subsection{End-to-end Training}
Existing self-supervised strategies \cite{ouasfi2024unsupervised,boulch2021needrop} struggle to accurately predict implicit fields from sparse point clouds. Here, we propose an effective framework to train our methods in an end-to-end manner. We first use BSP to map the sparse point cloud $Q$ into a continuous parametric point cloud $S$, providing more precise supervision for GDO. Next, we leverage the neural network $g$ to learn grid deformations to predict the implicit field. To further enhance the smoothness of the implicit field, GDO learns more consistent deformation directions from the gradients to improve overall details. Experimental results in the following validate the effectiveness of our method.
\section{Experiments}
\label{sec:experiments}

%With learned signed distances from each shape, we employ Marching Tetrahedra to reconstruct surface. All of experiments are conducted on a single GeForce RTX-3090 GPU. 

\begin{figure*}[!t]
  \centering
  \includegraphics[width=\linewidth]{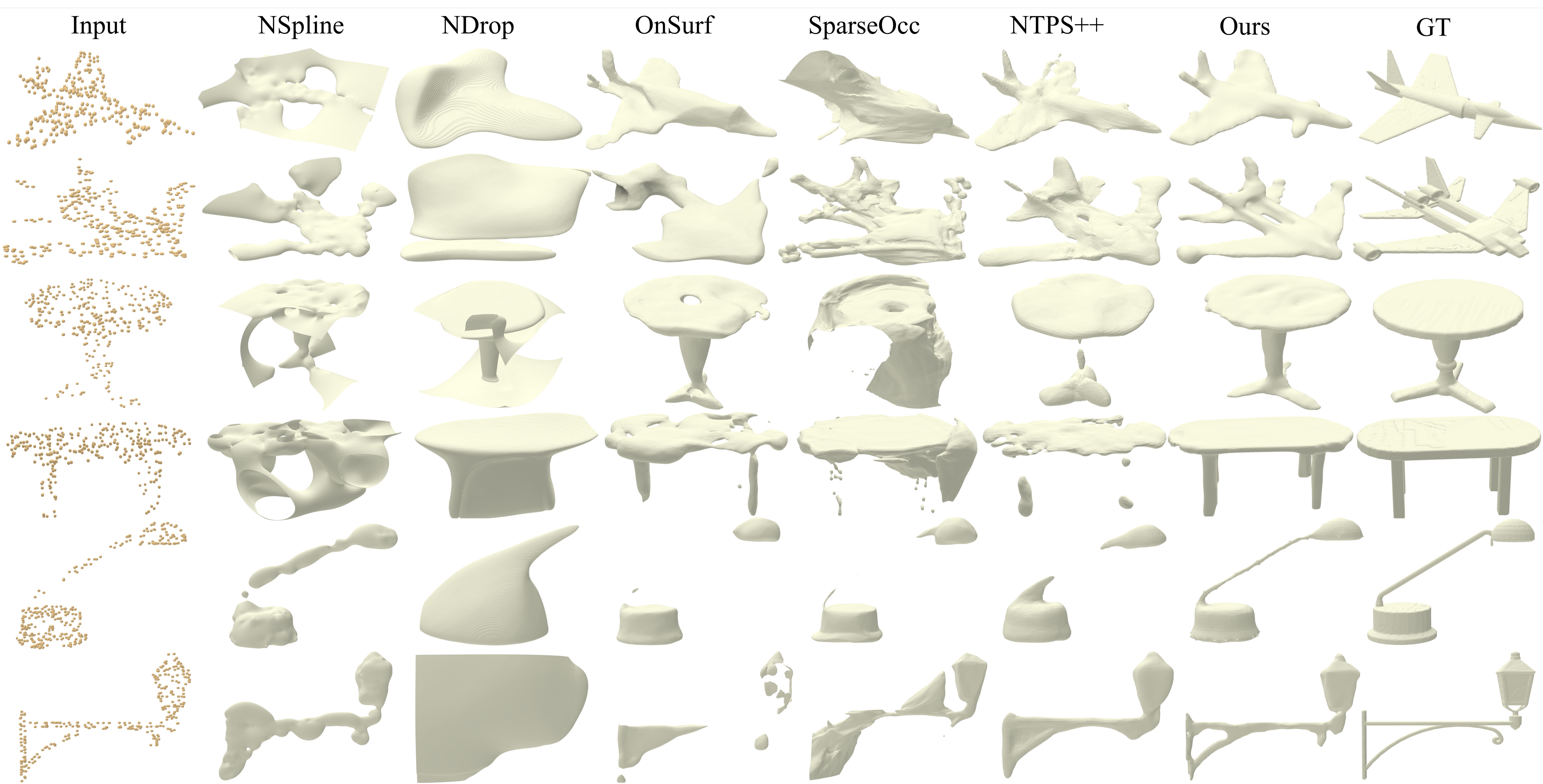}
  \caption{Visual comparison on ShapeNet. The input contains 300 points.}
  \label{fig:shapnet}
\end{figure*}

% \subsection{Baselines.} We compare our method with the state-of-art methods, including PSR \cite{kazhdan2013screened}, NDrop \cite{boulch2021needrop}, NP \cite{ma2021neural}, SAP \cite{peng2021shape}, Nspline \cite{williams2021neural}, VIPSS \cite{huang2019variational}, NTPS \cite{chen2023unsupervised}, Onsurf \cite{On-SurfacePriors}, SparseOcc \cite{ouasfi2024unsupervised} and IMLS \cite{wang2021neural}. In order to ensure the fairness, we conduct experiments according to the published code and default settings.

\begin{table}[!t]
  \centering
  \resizebox{1.0\linewidth}{!}{
  \begin{tabular}{c|c|c|c|c|c|c|c|c}
    \hline
     Class&Nspline& NP& NDrop&Onsurf&SparseOcc&NTPS&NTPS++&Ours \\
    \hline
    Plane &0.119&0.141&0.499&0.153& 0.219&0.095&0.088&\textbf{0.072}\\
    Chair &0.306&0.196&0.395&0.316& 0.183&0.197&0.195&\textbf{0.142}\\
    Cabinet &0.181&0.163&0.229&0.244& 0.220&0.138&0.137&\textbf{0.105}\\
    Display &0.193&0.145&0.287&0.204& 0.091&0.127&0.122&\textbf{0.099}\\
    Vessel &0.134&0.116&0.488&0.128& 0.158&0.104&0.101&\textbf{0.080}\\
    Table &0.318&0.400&0.426&0.288& 0.261&0.225&0.215&\textbf{0.108}\\
    Lamp &0.213&0.162&0.554&0.229& 0.192&0.120&0.112&\textbf{0.077}\\
    Sofa &0.168&0.139&0.259&0.147& 0.178&0.125&0.129&\textbf{0.116}\\
    \hline
    Mean &0.206&0.183&0.392&0.214& 0.187&0.141&0.137&\textbf{0.099}\\
    \hline
  \end{tabular}}
  \vspace{-0.30cm}
  \caption{Reconstruction accuracy under ShapeNet in terms of CD$_{L1}$ $\times$ 10.}
    \vspace{-0.30cm}
  \label{tab:shapenet-L1CD}
\end{table}
\begin{table}[!h]
  \centering
  \resizebox{1.0\linewidth}{!}{
  \begin{tabular}{c|c|c|c|c|c|c|c|c}
    \hline
     Class&Nspline& NP& NDrop&Onsurf& SparseOcc&NTPS&NTPS++&Ours \\
    \hline
    Plane &0.127&0.036&0.755&0.112& 0.165&0.030&0.026&\textbf{0.022}\\
    Chair &0.247&0.174&0.532&0.448& 0.162&0.149&0.140&\textbf{0.115}\\
    Cabinet &0.064&0.086&0.245&0.171& 0.178&0.050&0.050&\textbf{0.046}\\
    Display &0.095&0.099&0.401&0.153& 0.081&0.083&\textbf{0.078}&\textbf{0.078}\\
    Vessel &0.066&0.074&0.844&0.066& 0.073&0.051&0.046&\textbf{0.042}\\
    Table &0.312&0.892&0.701&0.419& 0.415&0.272&0.264&\textbf{0.188}\\
    Lamp &0.183&0.144&1.071&0.351& 0.466&0.051&0.047&\textbf{0.043}\\
    Sofa &0.053&0.072&0.463&0.066& 0.010&0.056&0.062&\textbf{0.052}\\
    \hline
    Mean &0.143&0.197&0.627&0.223& 0.193&0.093&0.089&\textbf{0.073}\\
    \hline
  \end{tabular}}
  \vspace{-0.10in}
  \caption{Reconstruction accuracy under ShapeNet in terms of CD$_{L2}$ $\times$ 100.}
  \label{tab:shapenet-L2CD}
\end{table}
\begin{table}[!h]
  \centering
  \resizebox{1.0\linewidth}{!}{
  \begin{tabular}{c|c|c|c|c|c|c|c|c}
    \hline
     Class&Nspline& NP& NDrop&Onsurf&SparseOcc&NTPS &NTPS++&Ours \\
    \hline
    Plane &0.895&0.897&0.819&0.864&0.853&0.899 &0.912&\textbf{0.913}\\
    Chair &0.759&0.861&0.777&0.813&0.844&0.863 &0.873&\textbf{0.896}\\
    Cabinet &0.840&0.888&0.843&0.787&0.813&0.898 &0.897&\textbf{0.904}\\
    Display &0.830&0.909&0.873&0.855&0.872&0.924 &\textbf{0.936}&0.927\\
    Vessel &0.842&0.880&0.838&0.879&0.841&0.908 &\textbf{0.913}&0.911\\
    Table &0.771&0.835&0.795&0.827&0.856&0.877 &0.888&\textbf{0.890}\\
    Lamp &0.814&0.887&0.828&0.858&0.883&0.902 &0.910&\textbf{0.914}\\
    Sofa &0.828&0.905&0.808&0.881&0.870&0.919 &0.915&\textbf{0.923}\\
    \hline
    Mean &0.822&0.883&0.823&0.845&0.854&0.899 &0.905&\textbf{0.909}\\
    \hline
  \end{tabular}}
  \vspace{-0.10in}
  \caption{Reconstruction accuracy under ShapeNet in terms of NC.}
  \label{tab:shapenet-NC}
  %\vspace*{-0.5cm}
\end{table}
\subsection{Experiment Setup}
\textbf{Datasets and Metrics.} We adopt five synthetic and real scanned datasets to evaluate our method. We first compare the performance of our method on D-FAUST \cite{bogo2017dynamic}, SRB \cite{williams2019deep}, and ShapeNet \cite{shapeneturl}, following the Ndrop and NTPS. To verify the applicability of the method under extremely sparse conditions, we follow NTPS to randomly sample 300 points for each shape as the input for ShapeNet and D-FAUST. %For fair comparison, we trained SparseOcc according to the default settings with open source code.%
For SRB dataset, we follow SparseOcc \cite{ouasfi2024unsupervised} to sample 1024 points for comparison. To further validate the effectiveness of in real large-scale scenarios, we validate our method on the 3DScene \cite{zhou2013dense} and KITTI \cite{geiger2012we}. For the 3DScene dataset, we follow previous methods to randomly sample 100 points $/m^{2}$. For the KITTI dataset, we use point clouds in single frames to conduct a comparison. 

We leverage L1 and L2 Chamfer Distance (CD$_{L1}$, CD$_{L2}$), Normal Consistency (NC) and Hausdorff Distance (HD) as evaluation metrics. For the shape and scene surface reconstruction, we sample $100k$ and $1000k$ points from the reconstructed and ground truth surfaces to calculate the distance errors.
\subsection{Surface Reconstruction On Shapes}
\textbf{ShapeNet.} We compare our method with Nspline\cite{williams2021neural}, NP\cite{ma2021neural}, NDrop, Onsurf, SparseOcc, NTPS++ and NTPS. The comparison results for different metrics are reported in Tab. \ref{tab:shapenet-L1CD}, Tab. \ref{tab:shapenet-L2CD} and Tab. \ref{tab:shapenet-NC}, where our method achieves the best results across all classes. We further present the visual comparison in Fig. \ref{fig:shapnet}. Ndrop and Nspline fail to generate accurate shape surfaces from sparse input, while NTPS++ and Onsurf generate correct shapes but with larger errors. SparseOcc cannot rely on decision boundaries to accurately predict occupancy fields under extremely sparse input conditions, making it challenging to reconstruct complex geometries. In contrast, our method produces more complete and smoother surfaces.
\begin{figure}[!t]
  \centering
  \includegraphics[width=0.9\linewidth]{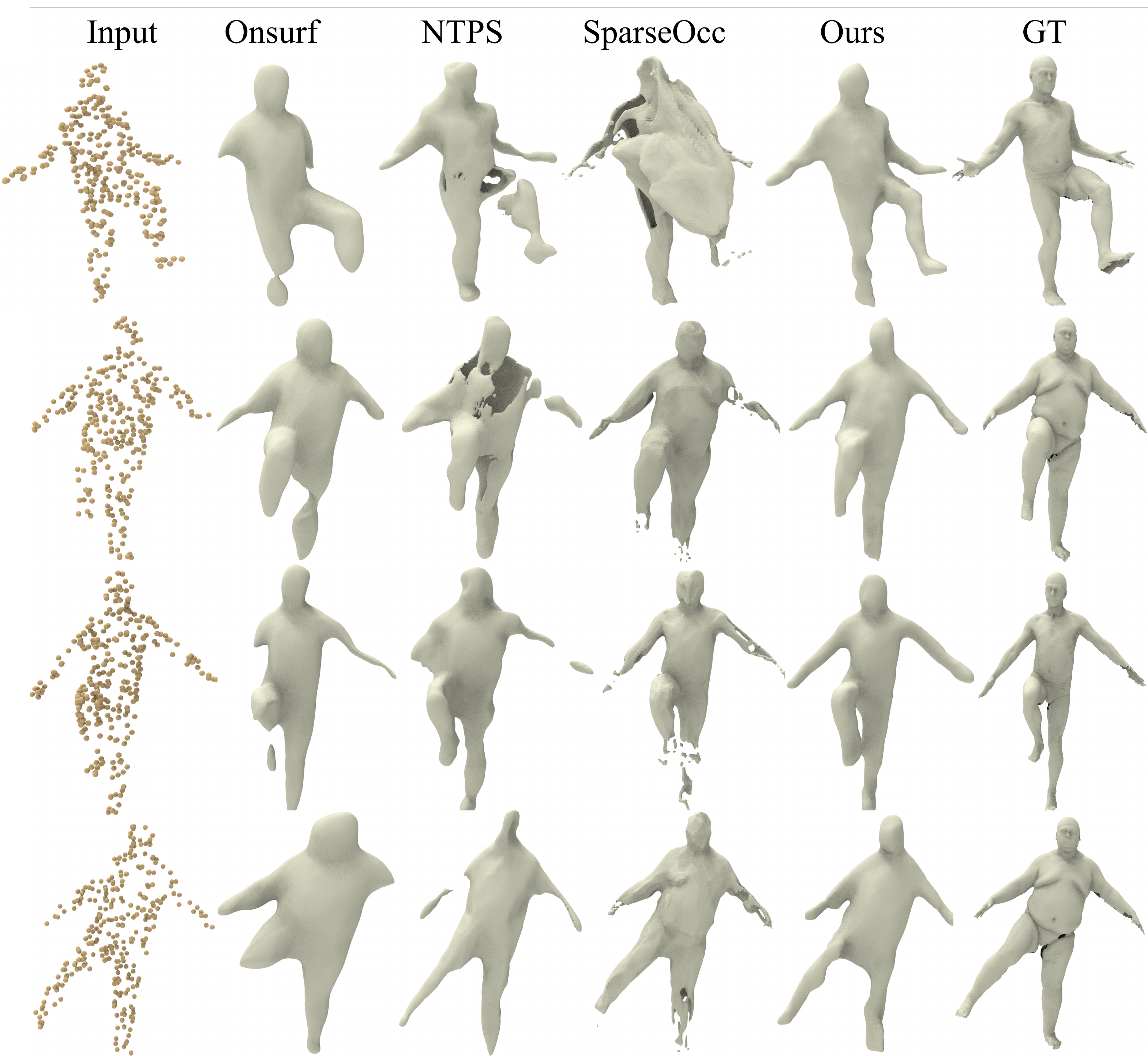}
  \caption{Visual comparison on D-FAUST. The input contains 300 points.}
  \label{fig:dfaust}
  \vspace*{-0.2cm}
\end{figure}
\begin{figure}[!t]
  \centering
  \includegraphics[width=\linewidth]{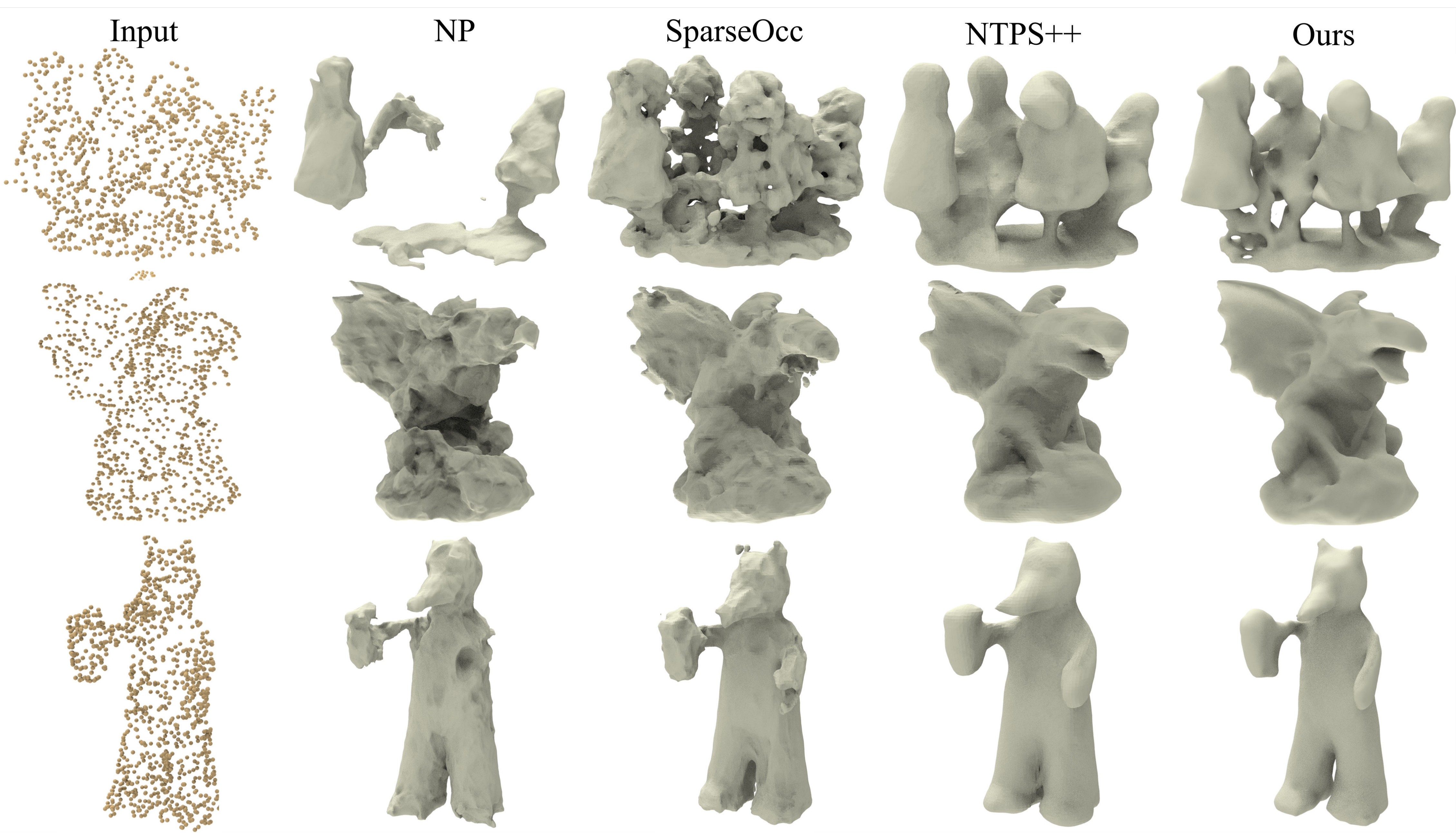}
  \caption{Visual comparison on SRB. The input contains 1024 points.}
  \label{fig:srb}
  \vspace*{-0.5cm}
\end{figure}

\begin{figure*}[!t]
  \centering
  \includegraphics[width=\linewidth]{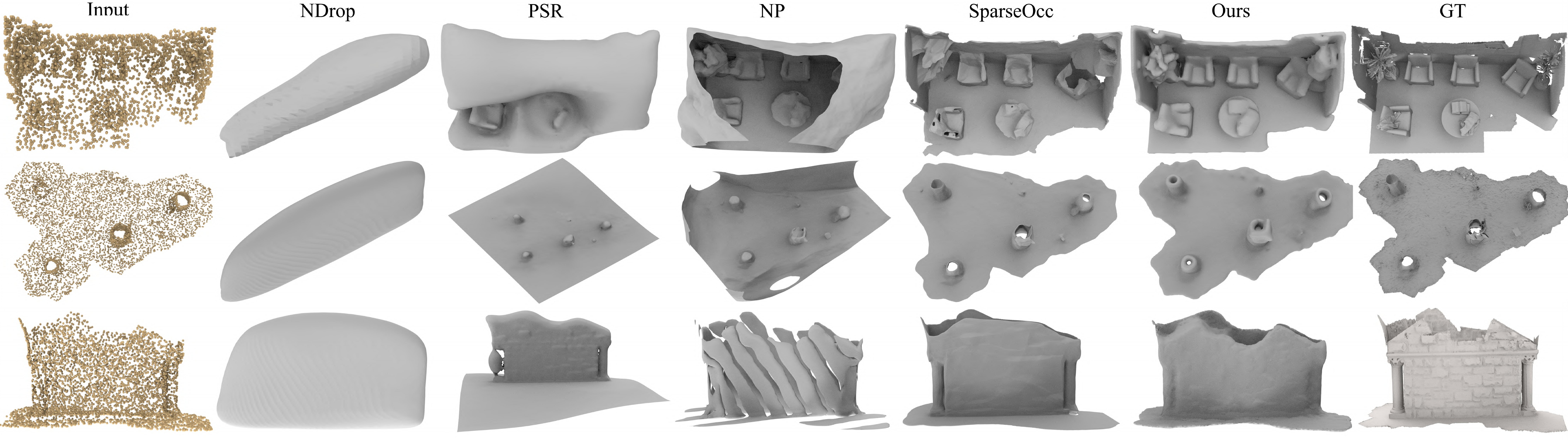}
  \caption{Visual comparison on 3DScene. The input contains $100$ points $/m^{2}$.}
  \label{fig:3dscene}
  \vspace*{-0.2cm}
\end{figure*}

\noindent \textbf{DFAUST.} As shown in Tab. \ref{tab:dfaust}, we follow Ndrop to report the 5\%, 50\% and 95\% of CD$_{L1}$, CD$_{L2}$, and NC results on the DFAUST dataset, achieving the best performance across all metrics. Additionally, we present a visual comparison with Onsurf, SparseOcc and NTPS in Fig. \ref{fig:dfaust}. Our method generates more complete human body with different poses.

\begin{figure*}
  \centering
  \includegraphics[width=0.9\linewidth]{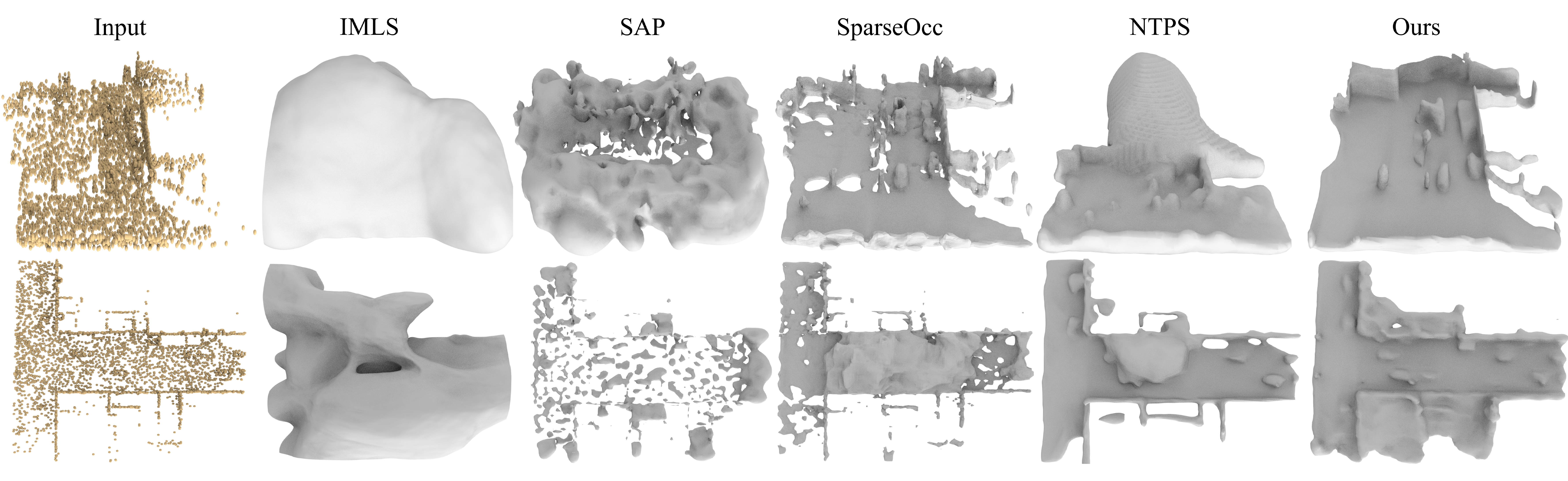}
  \caption{Visual comparison on KITTI-street.}
  \label{fig:kitti-street}
   \vspace{-0.2cm}
\end{figure*}

\begin{table}[!t]
\centering
\resizebox{0.7\linewidth}{!}{
    \begin{tabular}{c|c|c|c|c}
    \hline
    \multirow{2}{*}{Methods} & \multicolumn{3}{c|}{CD$_{L2} \times 100$} & \multirow{2}{*}{NC}\\
    \cline{2-4}
     & 5\% & 50\% & 95\% & \\
    \hline
    VIPSS & 0.518 & 4.327 & 9.383 & 0.890 \\
    NDrop & 0.126 & 1.000 & 7.404 & 0.792 \\
    NP & 0.018 & 0.032 & 0.283 & 0.877 \\
    Nspline & 0.037 & 0.080 & 0.368 & 0.808 \\
    SAP & 0.014 & 0.024 & 0.071 & 0.852 \\
    SparseOcc &0.012  &0.019  &\textbf{0.034}  & 0.870 \\
    OnSurf & 0.015 & 0.037 & 0.123 & 0.908 \\
    NTPS & 0.012 & 0.160 & 0.022 & 0.909 \\
    \hline
    Ours &\textbf{0.007} &\textbf{0.019} &{0.133} &\textbf{0.914}  \\
    \hline
    \end{tabular}}
\caption{Reconstruction accuracy under DFAUST in terms of CD$_{L2}$ $\times$ 100 and NC.}
\label{tab:dfaust}
\end{table}
\noindent\textbf{SRB.} We report the results on the real scanned dataset SRB in Tab. \ref{tab:srb} and present a visual comparison in Fig. \ref{fig:srb}. All baseline methods reconstruct coarse surfaces with input of 1024 points. In contrast, our method not only reconstructs the complete shape but also recovers more local details.

\begin{table}[!t]
  \centering
  \resizebox{0.6\linewidth}{!}{
  \begin{tabular}{c|c|c}
    \hline
     Methods&CD$_{L1}     \times 100$& HD\\
    \hline
    PSR &2.27&21.1\\
    NTPS&0.73&7.78\\
    NP &0.58&8.90\\
    NTPS++&0.66&7.30\\
    SparseOcc&0.49&6.04\\

    \hline
    Ours &\textbf{0.41} &\textbf{5.66}\\
    \hline
  \end{tabular}}
  \vspace{-0.10in}
  \caption{Reconstruction accuracy under SRB in terms of CD$_{L1}$ $\times$ 100 and HD.}
  \vspace{-0.5cm}
  \label{tab:srb}
\end{table}
\subsection{Surface Reconstruction On Scenes}
\textbf{3DScene.} We compare our method with the current state-of-the-art methods, including PSR\cite{kazhdan2013screened}, NP, Ndrop, SparseOcc, NTPS on the 3DScene dataset. The extensive results presented in Tab. \ref{tab:3dscene} demonstrate that our method performs well in real-world scenarios. As shown in Fig. \ref{fig:3dscene}, our method reconstructs smoother surfaces and captures more internal details than NP and SparseOcc.

\begin{figure}[!t]
  \centering
  \includegraphics[width=0.9\linewidth]{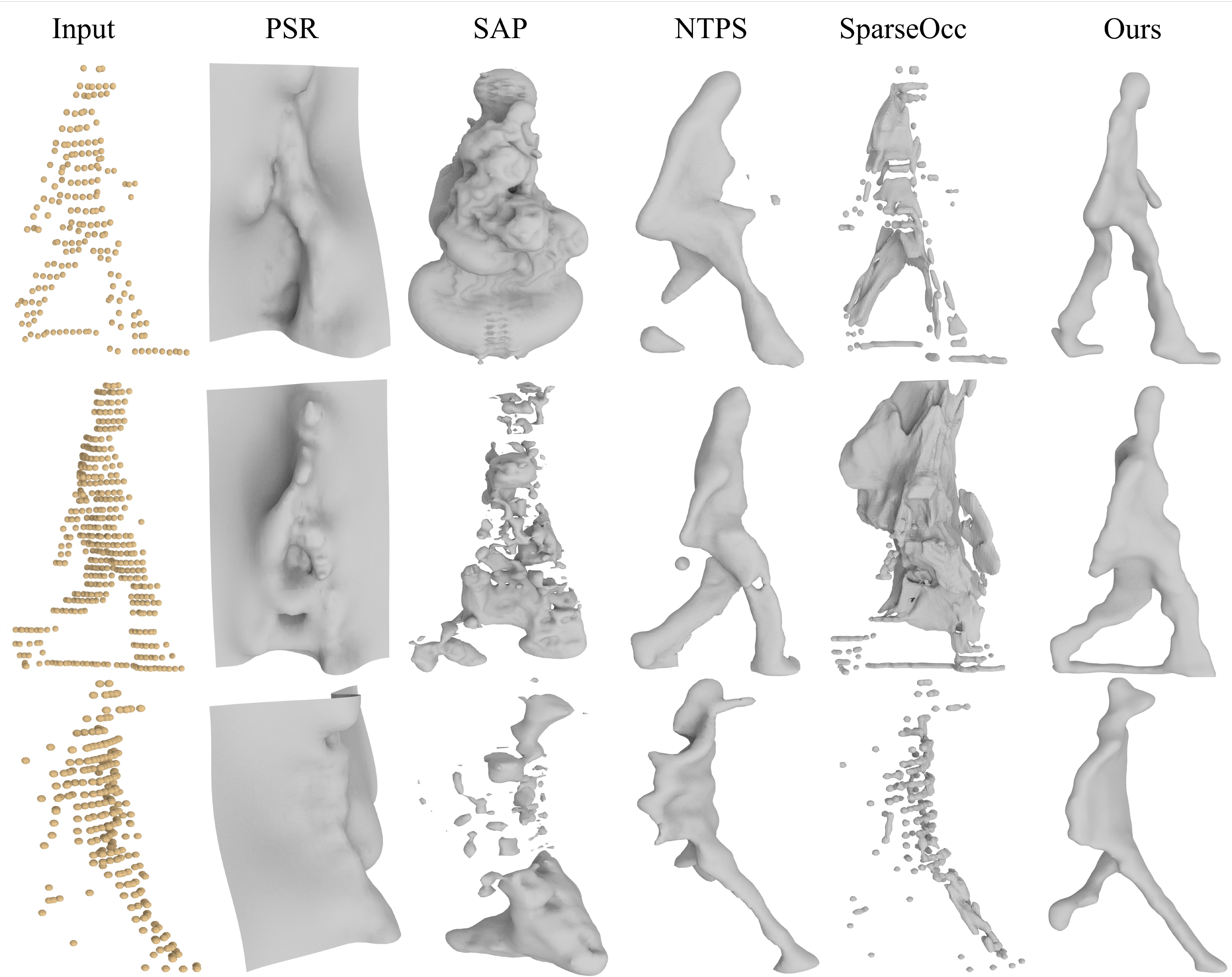}
  \caption{Visual comparison under KITTI-pedestrians.}
  \label{fig:kitti-human}
  \vspace*{-0.4cm}
\end{figure}

\begin{table}[!t]
  \centering
  \resizebox{1.0\linewidth}{!}{
    \begin{tabular}{c|c|c|c|c|c|c|c}
     \hline
     && PSR & NP & Ndrop & NTPS& SparseOcc& Ours \\
     \hline
    {Burghers}&CD$_{L1}$&0.178&0.064&0.200&0.055&0.022&\textbf{0.015}
    \\
    &CD$_{L2}$&0.205&0.008&0.114&0.005&\textbf{0.001}&\textbf{0.001}\\
    &NC&0.874&0.898&0.825&0.909&0.871&\textbf{0.890}\\
    \hline
    {Copyroom}&CD$_{L1}$&0.225&0.049&0.168&0.045&0.041&\textbf{0.037}
    \\
    &CD$_{L2}$&0.286&0.005&0.063&0.003&0.012&\textbf{0.003}\\
    &NC&0.861&0.828&0.696&0.892&0.812&\textbf{0.897}\\
    \hline
    {Lounge}&CD$_{L1}$&0.280&0.133&0.156&0.129&0.021&\textbf{0.012}\\
    &CD$_{L2}$&0.365&0.038&0.050&0.022&\textbf{0.001}&\textbf{0.001}\\
    &NC&0.869&0.847&0.663&0.872&0.870&\textbf{0.903}\\
    \hline
    {Stonewall}&CD$_{L1}$&0.300&0.060&0.150&0.054&0.028&\textbf{0.021}
    \\
    &CD$_{L2}$&0.480&0.005&0.081&0.004&0.003&\textbf{0.002}\\
    &NC&0.866&0.910&0.815&0.939&0.931&\textbf{0.937}\\
    \hline
    {Totempole}&CD$_{L1}$&0.588&0.178&0.203&0.103&0.026&\textbf{0.022}
    \\
    &CD$_{L2}$&1.673&0.024&0.139&0.017&\textbf{0.001}&\textbf{0.001}\\
    &NC&0.879&0.908&0.844&0.935&\textbf{0.936}&0.931\\
    \hline
    \end{tabular}
  }
  \vspace{-0.10in}
  \caption{CD$_{L1}$, CD$_{L2}$ and NC comparison under 3DScene.}
  \label{tab:3dscene}
  \vspace{-0.30in}
\end{table}
\noindent \textbf{KITTI.} We show a visual comparison with IMLS\cite{wang2021neural}, Ndrop, PSR, SAP, NTPS and SparseOcc on real scanned large-scale street and local pedestrians on the KITTI dataset in Fig. \ref{fig:kitti-street} and Fig. \ref{fig:kitti-human}. Decision boundary based methods like SparseOcc only capture cause shapes. In contrast, parameterized methods excel at reconstructing continuous surfaces. Our method can reconstruct more complete and detailed surfaces, such as diverse human poses and complex street scenes. 

% \begin{figure*}
%   \centering
%   \includegraphics[width=0.9\linewidth]{sec/pictures/kitti-street.pdf}
%   \caption{Visual comparison under KITTI-street.}
%   \label{fig:kitti-street}
%    \vspace{-0.4cm}
% \end{figure*}
\section{Ablation Studies}
\label{sec:ablation_studies}
\begin{figure}[!h]
  \centering
  \includegraphics[width=0.95\linewidth]{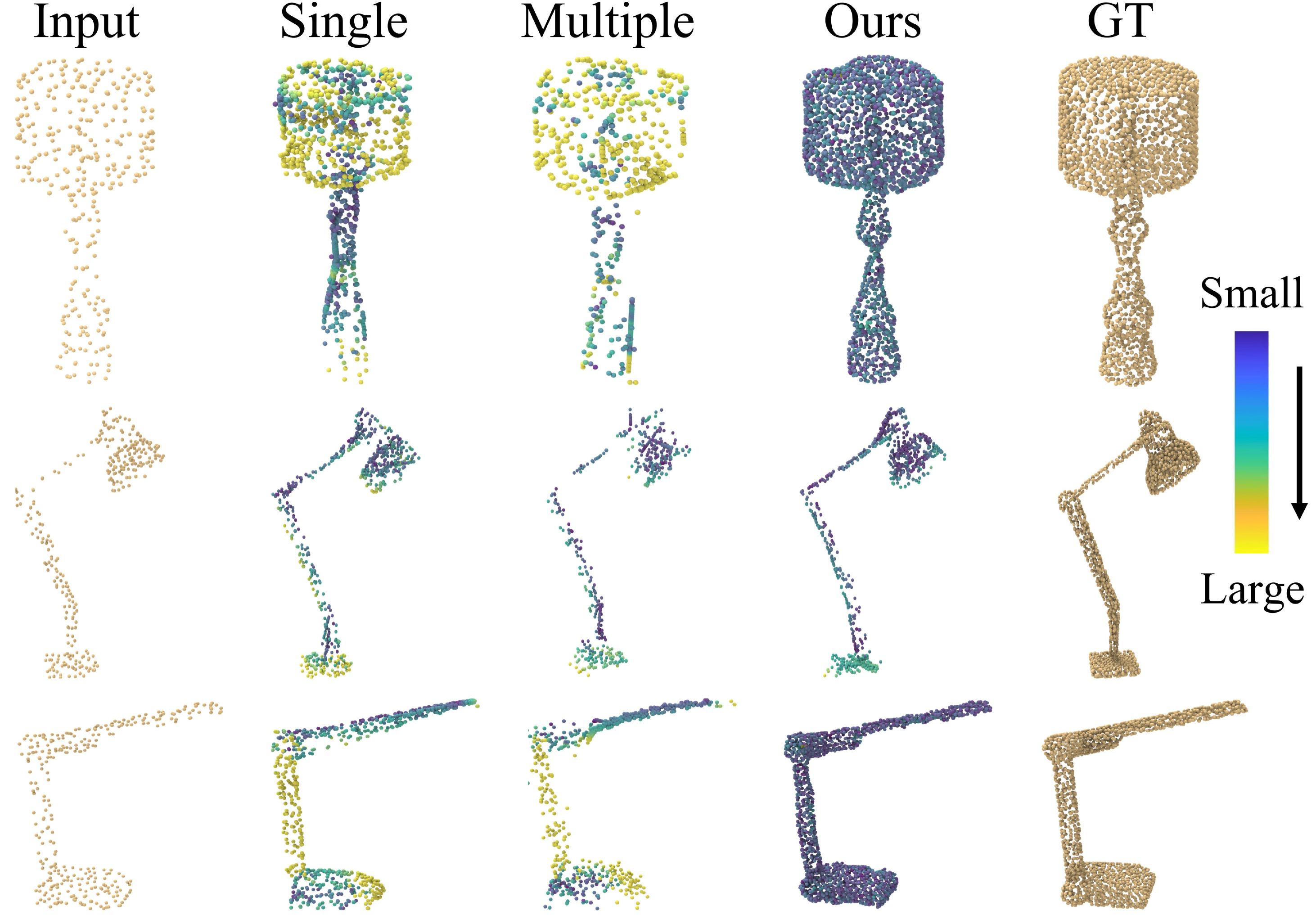}
  \caption{Effect of BSP. The color indicate the point distance error to ground truth surface.}
  \label{fig:ab_1}
  \vspace*{-0.8cm}
\end{figure}

To validate the effectiveness of each module, we conduct ablation experiments on the lamp class of ShapeNet dataset. We present the quantitative results and visualization under different experimental settings.\\
\textbf{Effect of BSP.} To evaluate the effectiveness of the BSP, we firstly remove the BSP and only rely on sparse input to infer signed distance functions (denoted as Sparse), which lead a significant increase in CD error. It indicates that the parameterized supervision generated by BSP has a substantial impact on reconstruction accuracy. Next, we replace the BSP with the parameterization strategies proposed by TPS \cite{chen2023unsupervised} and Atlas \cite{yu2022part}, which denoted as Single and Multiple, respectively. \\
As shown in Tab. \ref{tab:ab_bsp}, both Single and Multiple lead to an increase in CD error. We additionally compared the CD error maps of point clouds predicted by different parameterization methods in Fig. \ref{fig:ab_1} . Notably, single based parameterization (such as TPS) only generate a  coarse global surface. Meanwhile, the multi-part parameterization strategy based on AtlasNet exhibits truncation and overfitting in local regions. In contrast, BSP efficiently integrates local parameterized surfaces to construct a continuous global surface, achieving the best performance. To further illustrate the applicability of BSP to sparse input, we replace BSP with the state-of-art upsampling method LDI \cite{li2024learning} noted as Upsample. As shown in Tab. \ref{tab:ab_bsp}, LDI also struggles to predict accurate result due to the highly sparse distribution. We further provide detailed visualization comparison in the supplementary.
\vspace{-0.10in}
\begin{table}[!h]
  \centering
  \resizebox{0.7\linewidth}{!}{
  \begin{tabular}{c|c|c|c}
    \hline
     &CD$_{L1} \times 10$& CD$_{L2} \times 100$ &NC\\
    \hline
    Sparse &0.873&4.315&0.814\\
    Upsample&0.427&0.986&0.830\\
    Single &0.083&0.049&0.906\\
    Multiple&0.087&0.051&0.901\\
    %w/o norm &0.081&0.046&0.911\\
    \hline
    Ours &\textbf{0.077}&\textbf{0.043}&\textbf{0.914}\\
    \hline
  \end{tabular}}
  \vspace{-0.10in}
  \caption{Effect of BSP.}
  \vspace{-0.30cm}
  \label{tab:ab_bsp}
\end{table}

\begin{figure}[!h]
  \centering
  \includegraphics[width=0.95\linewidth]{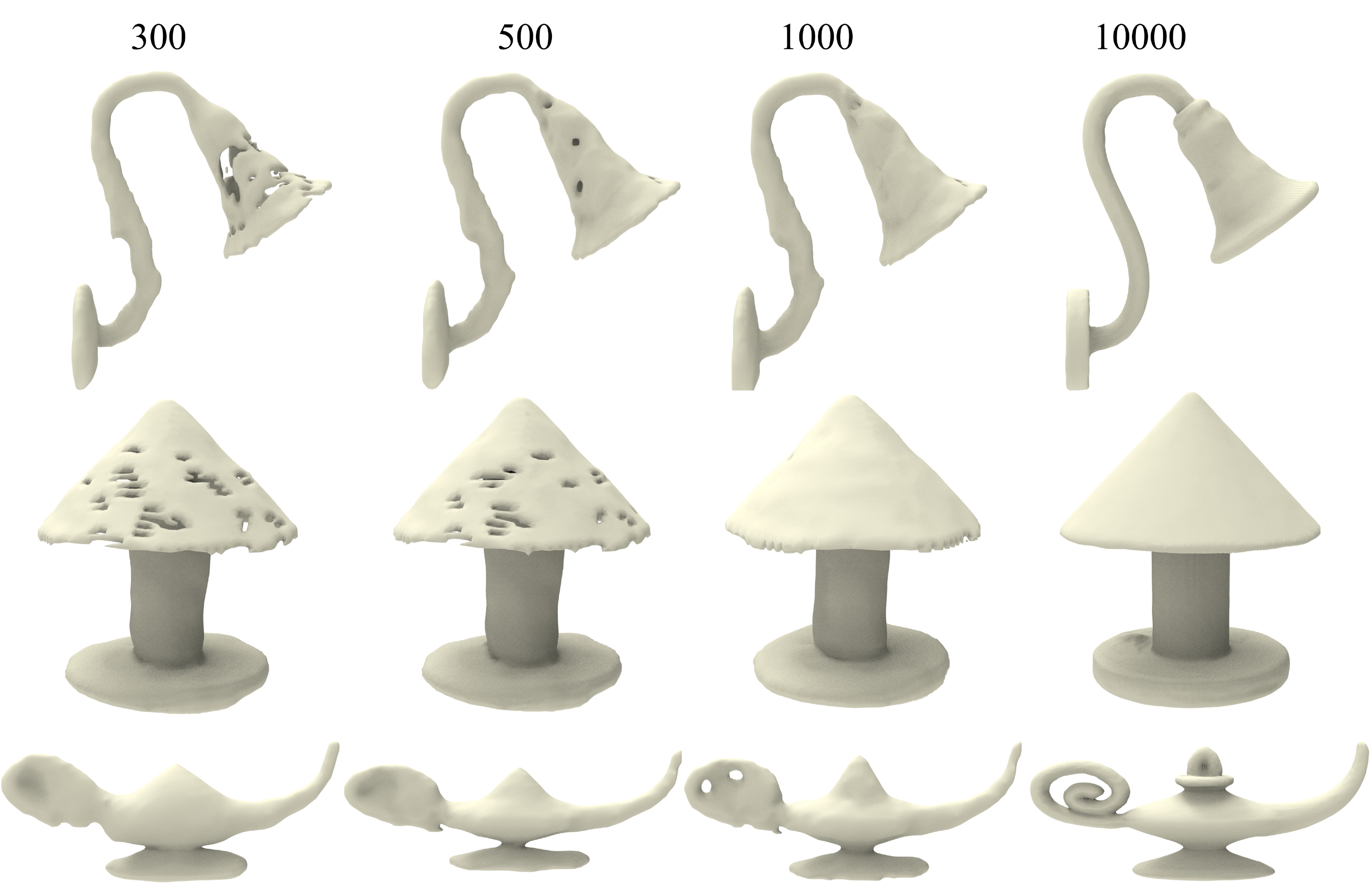}
  \caption{Visual comparison with levels of input size.}
    %\vspace{-0.20cm}
  \label{fig:ab_2}
\end{figure}

\noindent\textbf{Level of Input Size.} We evaluate the robustness of our method with different point size levels. Our visualization results are reported in Fig. \ref{fig:ab_2}. As the number of point clouds increases, we are able to generate more uniform parameterized surfaces and accurate geometries.\\
\noindent\textbf{Effect of GDO.} We demonstrate that GDO can learn a consistent deformation direction from gradients in Fig. \ref{fig:gdo}. Here, we further justify the effectiveness of GDO in inferring the implicit function $f$. We first remove the gradient consistency constrain, and only learn the implicit functions by predicting the grid point offsets, denoted as $f_{offset}$. Then, we remove GDO and apply TPS optimization strategy as baseline denoted as $f_{TPS}$. Both of them cause increases the CD error at different levels. As shown in Tab. \ref{tab:ab_gdo}, grid deformation-based strategies ($f_{offset}$ and Ours) achieve higher accuracy, and $f_{GDO}$ provides the most precise geometric surface prediction. We provide visualizations of reconstruction results under different optimization strategies in the supplementary. 
\vspace{-0.10in}
\begin{table}[!h]
  \centering
  \resizebox{0.7\linewidth}{!}{
  \begin{tabular}{c|c|c|c}
    \hline
     &CD$_{L1} \times 10$& CD$_{L2} \times 100$ &NC\\
    \hline
    $f_{TPS}$ &0.094&0.058&0.861\\
    $f_{offset}$ &0.089&0.055&0.883\\
    \hline
    Ours &\textbf{0.077}&\textbf{0.043}&\textbf{0.914}\\
    \hline
  \end{tabular}}
  \vspace{-0.3cm}
  \caption{Effect of GDO.}
  \label{tab:ab_gdo}
\end{table}
\vspace{-0.3cm}

\noindent\textbf{Effect of Loss Functions.} To validate the effectiveness of each optimization term, we present the results of removing different loss terms in Tab. \ref{tab:ab_loss} to assess their importance in our method. We first remove $L_{para}$ and rely solely on the sparse point cloud for reconstruction, which leads to a significant increase in CD error. To remove $L_{deform}$, we utilize pretrained BSP to obtain a parameterized surface as supervision without further optimization, which also results in decreased accuracy. Finally, we remove $L_{surf}$ results in slightly worse results. Overall, $L_{para}$ and $L_{deform}$ have a greater impact on the metrics, indicating that dense and further optimizable parameterized surface are important for learning accurate implicit functions.\\
\textbf{Number of KNN Samples.} We explore the effectiveness of different numbers of sampled single local patches in Tab. \ref{tab:sup_samples}. With the increasing of samples, the network can predict the parametric surface more precisely. However, when the hyper-parameter is set to 15, the improvements in accuracy become marginal. To consider the balance between performance and efficiency, we set this hyper-parameter to 10 by default.
\begin{table}[!h]
  \centering
    \vspace{-0.1cm}
  \resizebox{0.7\linewidth}{!}{
    \begin{tabular}{c|c|c|c}
      \hline
      & CD$_{L1} \times 10$ & CD$_{L2} \times 100$ & NC \\
      \hline
      w/o $L_{para}$ & 0.873 & 4.315 & 0.814 \\
      w/o $L_{deform}$ & 0.085 & 0.053 & 0.898 \\
      w/o $L_{surf}$ & 0.081 & 0.044 & 0.908 \\
      \hline
      Ours & \textbf{0.077} & \textbf{0.043} & \textbf{0.914} \\
      \hline
    \end{tabular}
  }
  \vspace{-0.10in}
  \caption{Effect of loss functions.}
  % \vspace{0.30cm}
  \label{tab:ab_loss}
\end{table}
\begin{table}[!h]
  \centering
  \vspace{-0.6cm}
  \resizebox{0.7\linewidth}{!}{% 调整表格宽度以适应页面
  \begin{tabular}{c|c|c|c|c} % 列分隔符
      \hline
     Sample Size & 3 & 5 & 10 & 15 \\
    \hline
    CD$_{L1} \times 10$ &0.086  &0.081  &0.077  &\textbf{0.075}  \\
    CD$_{L2} \times 100$  &0.049  &0.047  &0.043  &\textbf{0.041}  \\
    NC &0.896  &0.905  &\textbf{0.914}  &\textbf{0.914}  \\ % 修正了NC列的加粗问题
    \hline
  \end{tabular}
  }
  \vspace{-0.10in}
  \caption{Number of KNN samples.} % 表格标题
  \label{tab:sup_samples} % 表格标签
  \vspace*{-0.6cm}
\end{table}
\section{Conclusion}
We propose an innovative training framework that learns smooth implicit fields from sparse point cloud inputs and reconstructs complete and continuous surfaces. Unlike previous methods, we parametrize local surfaces by learning bijective functions and integrate them into a global surface to ensure shape continuity. Experimental results demonstrate that the BSP strategy can generate more accurate parametrized surfaces. Additionally, we introduce a novel approach to apply deformation networks to sparse reconstruction tasks and propose GDO to further improve the accuracy of implicit field predictions. We validate the effectiveness of our method across extensive datasets and ablation studies. The results demonstrate our robustness for under varied conditions and settings.

% WARNING: do not forget to delete the supplementary pages from your submission 
%\input{sec/X_suppl}
\clearpage
\newpage
{
    \small
    \bibliographystyle{ieeenat_fullname}
    \bibliography{main}
}
\end{document}